\documentclass[11pt,a4paper]{article}
\usepackage{authblk}
\usepackage[hyperref]{acl2021}
\usepackage{times}
\usepackage{latexsym}

\usepackage{siunitx}
\usepackage{amsmath}
\usepackage{stmaryrd}
\usepackage{mathrsfs}
\usepackage{setspace}
\usepackage{standalone}
\usepackage{bm}
\usepackage{mathtools}
\usepackage{amssymb}
\usepackage{stfloats}

\usepackage{tikz}
\usetikzlibrary{fadings,calc,shapes.misc,arrows,decorations.pathmorphing,backgrounds,positioning,fit,petri}
\usetikzlibrary{arrows,decorations.pathmorphing,backgrounds,positioning,fit,petri,calc}
\usetikzlibrary{snakes}
\usetikzlibrary{decorations.pathreplacing,arrows.meta, shapes.misc, matrix, decorations.markings}
\usetikzlibrary{positioning,shapes,calc,shadows}
\usepackage{tikz-3dplot}
\newcommand*\concat{\mathbin{\|}}

  \DeclareMathOperator{\BERT}{BERT}
    \DeclareMathOperator{\GAT}{GAT}
      \DeclareMathOperator{\FFN}{FFN}

         \DeclareMathOperator{\cossim}{sim}
   \DeclareMathOperator{\total}{total}
    \DeclareMathOperator{\prior}{prior}
     \DeclareMathOperator{\task}{task}

\usepackage[english]{babel}
\usepackage{makecell}
\usepackage{booktabs}
\usepackage{graphicx, accents}
\newsavebox\mybox
\usepackage{subcaption}
\usepackage{mwe}
\usepackage{pifont}
\usepackage{soul}
\usepackage{color}

\definecolor{light-gray}{gray}{0.85}
\definecolor{dark-gray}{gray}{0.55}

\DeclareRobustCommand{\best}[1]{{\sethlcolor{dark-gray}\hl{#1}}}

\newcount\gaussF
\edef\gaussR{0}
\edef\gaussA{0}

\makeatletter
\pgfmathdeclarefunction{gaussR}{0}{%
 \global\advance\gaussF by 1\relax
 \ifodd\gaussF
  \pgfmathrnd@%
  \ifdim\pgfmathresult pt=0.0pt\relax%
    \def\pgfmathresult{0.00001}%
  \fi
  \pgfmathln@{\pgfmathresult}%
  \pgfmathmultiply@{-2}{\pgfmathresult}%
  \pgfmathsqrt@{\pgfmathresult}%
  \global\let\gaussR=\pgfmathresult
  \pgfmathrnd@%
  \pgfmathmultiply@{360}{\pgfmathresult}%
  \global\let\gaussA=\pgfmathresult
  \pgfmathcos@{\pgfmathresult}%
  \pgfmathmultiply@{\pgfmathresult}{\gaussR}%
 \else
  \pgfmathsin@{\gaussA}%
  \pgfmathmultiply@{\gaussR}{\pgfmathresult}%
 \fi
}

\pgfmathdeclarefunction{invgauss}{2}{%
  \pgfmathln{#1}
  \pgfmathmultiply@{\pgfmathresult}{-2}%
  \pgfmathsqrt@{\pgfmathresult}%
  \let\@radius=\pgfmathresult%
  \pgfmathmultiply{6.28318531}{#2}
  \pgfmathdeg@{\pgfmathresult}%
  \pgfmathcos@{\pgfmathresult}%
  \pgfmathmultiply@{\pgfmathresult}{\@radius}%
}

\pgfmathdeclarefunction{randnormal}{0}{%
  \pgfmathrnd@
  \ifdim\pgfmathresult pt=0.0pt\relax%
    \def\pgfmathresult{0.00001}%
  \fi%
  \let\@tmp=\pgfmathresult%
  \pgfmathrnd@%
  \ifdim\pgfmathresult pt=0.0pt\relax%
    \def\pgfmathresult{0.00001}%
  \fi
  \pgfmathinvgauss@{\pgfmathresult}{\@tmp}%
}

\usepackage{microtype}

\aclfinalcopy 

\title{Dynamic Contextualized Word Embeddings}

\author[*$\ddag$]{Valentin Hofmann}
\author[$\dag$*]{Janet B. Pierrehumbert}
\author[$\ddag$]{Hinrich Sch\"utze}

\affil[*]{Faculty of Linguistics, University of Oxford}
\affil[$\dag$]{Department of Engineering Science, University of Oxford}
\affil[$\ddag$]{Center for Information and Language Processing, LMU Munich \protect\\ \texttt{valentin.hofmann@ling-phil.ox.ac.uk}}

\date{}

\begin{document}

\maketitle

\begin{abstract}
Static word embeddings that represent words by a single vector 
 cannot capture
the variability of word meaning in different linguistic 
and extralinguistic contexts. Building on prior work on contextualized and dynamic
word
   embeddings, we introduce dynamic 
contextualized word embeddings 
that represent words as a function of both linguistic
and extralinguistic context.
Based on a pretrained language model (PLM), dynamic contextualized word embeddings model time and social space jointly, which 
makes them attractive for a range of NLP tasks involving 
semantic variability. We highlight potential application scenarios
by means of qualitative and quantitative
analyses on four English datasets.
\end{abstract}

\section{Introduction}

Over the last decade, word embeddings
have revolutionized the field of NLP. Traditional methods such as LSA \citep{Deerwester.1990}, 
word2vec \citep{Mikolov.2013b, Mikolov.2013c}, GloVe 
\citep{Pennington.2014}, and fastText \citep{Bojanowski.2017} compute \emph{static} word 
embeddings, i.e., they represent words as a single vector.
From a theoretical standpoint, this way of modeling lexical semantics
is problematic since it ignores 
the variability of word meaning in different linguistic contexts (e.g., polysemy)
as well as different extralinguistic contexts (e.g., temporal and social variation).

\begin{figure}[t!]
        \centering
        \tdplotsetmaincoords{70}{110}

\begin{tikzpicture}[scale=3,tdplot_main_coords]

\def\npoints{50}

\pgfmathsetseed{1}

\coordinate (O) at (0,0,0);

\draw[thick,->] (0,0,0) -- (0.7,0,0) node[anchor=north east]{};
\draw[thick,->] (0,0,0) -- (0,0.7,0) node[anchor=north west]{};
\draw[thick,->] (0,0,0) -- (0,0,0.7) node[anchor=south]{};

\pgfmathsetmacro{\ax}{0.2}
\pgfmathsetmacro{\ay}{-0.3}
\pgfmathsetmacro{\az}{0.6}

\pgfmathsetmacro{\bx}{0.2}
\pgfmathsetmacro{\by}{0.4}
\pgfmathsetmacro{\bz}{0.4}

\filldraw[fill=gray,opacity=0.2,draw=none] (O) -- (\ax,\ay,\az) -- (\bx,\by,\bz);

\tdplotdefinepoints(0,0,0)(\ax,\ay,\az)(\bx,\by,\bz)
\tdplotdrawpolytopearc{0.15}{anchor=north east, shift={(-0.1, 0)}}{$\phi_{ij}^{(k)}$}

\foreach \n in {0,1,...,\npoints}{
	\pgfmathsetmacro{\x}{gaussR/8}
	\pgfmathsetmacro{\y}{gaussR/8}
	\pgfmathsetmacro{\z}{gaussR/8}
	\path (\ax,\ay,\az) -- ++ (\x,\y,\z) node [circle,fill=blue!23,inner sep=0pt,minimum size=1mm] {};}

%

	
\node[circle, fill=blue!60,inner sep=0pt,minimum size=1mm,label={[label distance=-0.3cm]120:$\mathbf{e}_{ij}^{(k)}$}](e_k) at (\ax,\ay,\az) {};



	
\node[circle,fill=red!60,inner sep=0pt,minimum size=1mm,label={[label distance=-0.1cm]10:$\tilde{\mathbf{e}}^{(k)}$}] (e_k_tilde)at (\bx,\by,\bz) {};

\draw[->,bend left=45,shorten >=1pt] (e_k_tilde) to  node[below,near end] {$d$} (e_k);

\end{tikzpicture}
        \caption{Dynamic contextualized word embeddings. A static embedding $\tilde{\mathbf{e}}^{(k)}$ (\tikz[baseline=-0.6ex] \node[circle, fill=red!60,inner sep=0pt,minimum size=1.3mm] (0, 3) {};) is mapped to a dynamic embedding 
        $\mathbf{e}^{(k)}_{ij}$ (\tikz[baseline=-0.6ex] \node[circle, fill=blue!60,inner sep=0pt,minimum size=1.3mm] (0, 3) {};) by a function $d$ that takes time and social space into account. The 
        scattered points (\tikz[baseline=-0.6ex] \node[circle, fill=blue!23,inner sep=0pt,minimum size=1.3mm] (0, 3) {};) are contextualized versions of $\mathbf{e}^{(k)}_{ij}$. Variability in $\phi^{(k)}_{ij}$ indicates semantic dynamics across time and social space. The embeddings have 768 dimensions.} \label{fig:framework}
\end{figure}

The first shortcoming was addressed by 
the introduction of \emph{contextualized} word embeddings that
represent words as vectors varying across linguistic contexts. This allows them to capture more complex characteristics of word
meaning, including polysemy.
Contextualized
word embeddings are widely used in NLP, constituting the semantic backbone of pretrained language models (PLMs)
such as ELMo \citep{Peters.2018}, BERT \citep{Devlin.2019}, GPT-2 \citep{Radford.2019}, XLNet \citep{Yang.2019}, ELECTRA \citep{Clark.2020}, and T5 \citep{Raffel.2020}. 

A concurrent line of work focused on the second shortcoming of static word embeddings, resulting in various types of \emph{dynamic}
word embeddings. Dynamic word embeddings represent words as vectors varying across 
extralinguistic contexts, in particular time (e.g., \citealp{Rudolph.2018})
and social space (e.g., \citealp{Zeng.2018}).

In this paper, we introduce \emph{dynamic contextualized} word embeddings that combine the strengths of contextualized word embeddings with the flexibility
of dynamic word embeddings. Dynamic contextualized word embeddings mark a departure from existing contextualized word embeddings (which are not dynamic) as well as existing dynamic word embeddings (which are not contextualized). Furthermore, as opposed to all existing dynamic word embedding types, they represent time and social space jointly.

While our general framework for training dynamic contextualized word embeddings is model-agnostic (Figure \ref{fig:framework}),
 we present a version using a PLM (BERT) as the contextualizer,
 which allows for an easy integration within existing architectures.
Dynamic contextualized word embeddings can serve as an analytical tool (e.g., to track the emergence and spread of 
semantic changes in online communities) or be employed for downstream tasks (e.g.,
to build temporally and socially aware text classification models), making them beneficial
for various areas in NLP that face semantic variability.
 We illustrate application scenarios
by performing exploratory experiments on English data from ArXiv, Ciao, Reddit, and YELP.

\textbf{Contributions.} We introduce dynamic contextualized
word embeddings that represent words as a function of both linguistic
and extralinguistic context. Based on a PLM, dynamic contextualized word embeddings model time and social space jointly, which 
makes them attractive for a range of NLP tasks. We showcase potential applications by means of qualitative and quantitative
analyses.\footnote{We make our code publicly available at \url{https://github.com/valentinhofmann/dcwe}.}

\section{Related Work} \label{sec:related-work}

\subsection{Contextualized Word Embeddings}

The distinction
between the non-contextualized core meaning of a word and the 
senses that are realized in specific 
linguistic contexts lies at the heart of lexical-semantic scholarship \citep{Geeraerts.2010}, 
going back to at least \citet{Paul.1880}.
In NLP, this
is reflected
by contextualized word
embeddings that map type-level representations
to token-level representations as a function of the
linguistic context \citep{McCann.2017}.
As part of PLMs \citep{Peters.2018, Devlin.2019,  Radford.2019, Yang.2019, Clark.2020, Raffel.2020},
contextualized word embeddings have led to substantial
performance gains on a variety of tasks compared to static word embeddings
that 
only have type-level representations \citep{Deerwester.1990, Mikolov.2013b, Mikolov.2013c, Pennington.2014, Bojanowski.2017}.

Since their introduction, several studies have analyzed the 
linguistic properties of contextualized word embeddings
\citep{Peters.2018b, Goldberg.2019, Hewitt.2019, Jawahar.2019, Lin.2019, Liu.2019, Tenney.2019,
 Edmiston.2020, Ettinger.2020, Hofmann.2020c, Rogers.2020}.
Regarding lexical semantics, this line of research has 
shown that contextualized word embeddings are more context-specific
in the upper layers of a contextualizer \citep{Ethayarajh.2019, Mickus.2020, Vulic.2020} and represent different word senses as 
separated clusters \citep{Peters.2018, Coenen.2019, Wiedemann.2019}.

\subsection{Dynamic Word Embeddings} \label{sec:lit-dwe}

The meaning of a word can also vary across extralinguistic contexts such as 
time \citep{Bybee.2015, Koch.2016} and social space \citep{Robinson.2010, Robinson.2012, Geeraerts.2018}.
To capture these phenomena, various types of dynamic word embeddings have been proposed:
diachronic word embeddings for temporal semantic change \citep{Bamler.2017, Rosenfeld.2018, Rudolph.2018,Yao.2018, Gong.2020} and
personalized word embeddings for social semantic variation \citep{Zeng.2017, Zeng.2018, Oba.2019, Welch.2020, Welch.2020b, Yao.2020}. Other studies have demonstrated that
performance on a diverse set of tasks can be increased by including temporal \citep{Jaidka.2018, Lukes.2018} and social information \citep{Amir.2016, Hamilton.2016b, Yang.2016, Yang.2017, Hazarika.2018, Mishra.2018, delTredici.2019b, Li.2019, Mishra.2019}.

The relevance of dynamic (specifically diachronic) word embeddings is
also
reflected by the emergence of lexical semantic change detection 
as an established task in NLP \citep{Kutuzov.2018, Schlechtweg.2018,
Tahmasebi.2018, Dubossarsky.2019, Schlechtweg.2019, Asgari.2020, Pomsl.2020, Pravzak.2020, Schlechtweg.2020b, Schlechtweg.2020}.
 Besides dynamic word embeddings, many studies
 on lexical semantic change detection use methods based on static word embeddings \citep{Kim.2014, Kulkarni.2015}, e.g., 
the alignment of
  static word embedding spaces \citep{Hamilton.2016}. 
However, such approaches come at the cost of
modeling disadvantages \citep{Bamler.2017}.

Sociolinguistics has shown that temporal and social variation in language are tightly interwoven:  innovations such as a new word sense in the case of lexical semantics spread through 
the language community along social ties \citep{Milroy.1980, Milroy.1992, Labov.2001, Pierrehumbert.2012b}. 
However, most proposed dynamic word embedding types cannot capture more than one dimension of variation. Recently, a few studies 
 have taken first
steps in this direction by using genre information within a Bayesian model 
of semantic change \citep{Frermann.2016, Perrone.2019} and including social variables in
training diachronic word embeddings \citep{Jawahar.2019b}.
In addition, to capture
the full range of lexical-semantic variability,
dynamic word embeddings should also be contextualized. Crucially, while contextualized word 
embeddings have been used to investigate semantic change \citep{Giulianelli.2019, Hu.2019, Giulianelli.2020, Kutuzov.2020, Martinc.2020b, Martinc.2020}, the word embeddings employed in these studies are not dynamic, i.e., they represent a word in
a specific linguistic context by the same contextualized word
embedding independent of extralinguistic context or are fit to different time periods
as separate models.\footnote{It is interesting to notice that contextualized word embeddings so far have performed
worse than non-contextualized word embeddings on the task of lexical semantic change detection \citep{Kaiser.2020, Schlechtweg.2020}.}

\section{Model}

\subsection{Model Overview}

Given a sequence of words $X = \left[ x^{(1)}, \dots, x^{(K)} \right]$ 
and corresponding non-contextualized 
embeddings $E = \left[ \mathbf{e}^{(1)}, \dots, \mathbf{e}^{(K)} \right]$, 
contextualizing language models compute the
contextualized embedding of a particular word $x^{(k)}$, $\mathbf{h}^{(k)}$,
as a function $c$ of its non-contextualized embedding, $\mathbf{e}^{(k)}$,
and the non-contextualized embeddings 
of words in the left context $X^{(< k)}$ and the right context $X^{(> k)}$,\footnote{Some contextualizing language models such as GPT-2 \citep{Radford.2019} only operate on $X^{(< k)}$.}
\begin{equation} \label{eq:context}
\mathbf{h}^{(k)} = c \left( \mathbf{e}^{(k)}, E^{(< k)}, E^{(> k)}  \right).
\end{equation}
Crucially, while $\mathbf{h}^{(k)}$ is a token-level representation, $\mathbf{e}^{(k)}$
is a type-level representation and is modeled as a simple
embedding look-up.
Here, in order to take the variability of 
word meaning in different extralinguistic
contexts into account,
we depart from this practice
and model
$\mathbf{e}^{(k)}$ as a function $d$ that depends not only on the 
identity of $x^{(k)}$ but also on
the social context $s_i$ and the temporal context $t_j$ in which 
the sequence $X$ occurred,
\begin{equation} \label{eq:d}
\mathbf{e}^{(k)}_{ij} = d \left( x^{(k)}, s_i, t_j   \right).
\end{equation}
Dynamic contextualized word embeddings are hence computed in two stages: words are first mapped 
to dynamic type-level representations by $d$ and then to contextualized 
token-level representations by $c$ (Figures \ref{fig:framework} and \ref{fig:architecture}). This
two-stage structure follows work in cognitive science and linguistics
that indicates
that extralinguistic information is processed before linguistic information
by human speakers \citep{Hay.2006}.

\begin{figure}[t!]
        \centering
        \begin{tikzpicture}
[xscale=0.9,
 thick,
 model/.style={draw,rectangle,inner sep=6pt,rounded corners=6pt},
 sum/.style={draw,rectangle,inner sep=2pt,rounded corners=6pt},
 vec/.style={draw,rectangle,fill=white,inner sep=0pt,minimum width=1.5cm,minimum height=0.6cm,text height=0.3cm,text depth=0.05cm,rounded corners=8pt},
  token/.style={draw,rectangle,fill=white,inner sep=0pt,minimum width=0.65cm,minimum height=0.6cm,text height=0.3cm,text depth=0.05cm},
  comp/.style={draw,rectangle,fill=white,inner sep=0pt,minimum width=5cm,minimum height=1cm,text height=0.3cm,text depth=0.05cm},
 place/.style={circle,draw,fill=white,
inner sep=0pt,minimum size=5mm}]

\pgfmathsetseed{1} 

\foreach \x / \xtext in {0/$\tilde{\mathbf{e}}^{(1)}$,2/$\tilde{\mathbf{e}}^{(2)}$,4/$\tilde{\mathbf{e}}^{(3)}$}
	\node [vec] (tok-\x) at (\x + 0.375,1) {\xtext};
	
\foreach \x in {0,2,4}
	\node [] () at (\x + 0.375,0.5) {$+$};
	
\foreach \x / \xtext in {0/$\mathbf{o}_{ij}^{(1)}$,2/$\mathbf{o}_{ij}^{(2)}$,4/$\mathbf{o}_{ij}^{(3)}$}
	\node [vec] (o-\x) at (\x + 0.375,0) {\xtext};

	\foreach \pos / \x / \xtext in {-.1/0/$x^{(1)}$,0.9/1/$x^{(2)}$,1.9/2/$x^{(3)}$}
	\node [token] (x-\x) at (\pos ,-1.5) {\xtext};

\node[comp](bert)[above=0.7cm of tok-2, xshift=3mm,yshift=3mm] {$\BERT$};

\node[](loss)[above=0.7cm of bert] {$\mathcal{L}_{\task}$};

\node[vec,draw=none,fill=none](s_2_dummy_a)[below=0.7cm of o-4] {};
\node[vec,draw=none,fill=none](s_3_dummy_a)[below=1cm of o-4] {};

\node[token,draw=none,fill=none](s_1_dummy_b)[below=0.7cm of o-2] {};
\node[vec,draw=none,fill=none](s_2_dummy_b)[below=0.7cm of o-0] {};
\node[vec,draw=none,fill=none](s_3_dummy_b)[below=1cm of o-0] {};

\node[vec,draw=none,fill=none](s_3_dummy)[above=0.7cm of tok-0,xshift=3mm,yshift=3mm] {};
\node[vec,draw=none,fill=none](s_4_dummy)[above=0.7cm of tok-4,xshift=3mm,yshift=3mm] {};

\node[place](s_1)[below=0.65cm of o-4] {$s_i$};
\node[place](s_2)[left=0.42cm of s_1,yshift=-0.2cm] {};
\node[place](s_3)[below left=0.15cm and 0.05cm of s_1]  {};
\node[place](s_4) [below right=0.1cm and 0.12cm of s_1]  {};

	\path (s_1) edge[->,in=-90,out=90,dashed] ([yshift=-.234cm]o-0.south);
	\path (x-0) edge[->,in=180,out=180,dashed] ([xshift=-.257cm]o-0.west);
		\path (x-0) edge[->,in=180,out=180,dashed] ([xshift=-.257cm]tok-0.west);
		
	\path (bert) edge[->,dashed] (loss);
		
\path ([yshift=.075cm]tok-0.north) edge[->,in=-90,out=90,dashed] ([xshift=-1.2cm,yshift=-.234cm]bert.south);

\draw[](s_1) to (s_2);
\draw[](s_1) to (s_3);
\draw[](s_1) to (s_4);

\begin{scope}[on background layer]
\node ()[xshift=6mm,yshift=6mm,model,red!10,fill=red!1,very thick,fit=(x-2)(s_2_dummy_b) (s_3_dummy_b)] {};
\end{scope}

\begin{scope}[on background layer]
\node ()[xshift=4mm,yshift=4mm,model,red!23,fill=red!2.3,very thick,fit=(x-2)(s_2_dummy_b) (s_3_dummy_b)] {};
\end{scope}

\begin{scope}[on background layer]
\node ()[xshift=2mm,yshift=2mm,model,red!36,fill=red!3.6,very thick,fit=(x-2)(s_2_dummy_b) (s_3_dummy_b)] {};
\end{scope}

\begin{scope}[on background layer]
\node ()[model,red!50,fill=red!5,very thick,fit=(x-2)(s_2_dummy_b) (s_3_dummy_b)] {};
\end{scope}

\begin{scope}[on background layer]
\node (last) [xshift=6mm,yshift=6mm,model,red!10,fill=red!1,very thick,fit= (s_2_dummy_a)
(s_3_dummy_a) (s_2)  ] {};
\end{scope}

\begin{scope}[on background layer]
\node[xshift=4mm,yshift=4mm,model,red!23,fill=red!2.3,very thick,fit=(s_2_dummy_a)(s_3_dummy_a) (s_2)  ] {};
\end{scope}

\begin{scope}[on background layer]
\node[xshift=2mm,yshift=2mm,model,red!36,fill=red!3.6,very thick,fit= (s_2_dummy_a)(s_3_dummy_a) (s_2) ] {};
\end{scope}

\begin{scope}[on background layer]
\node (first)[model,red!50,fill=red!5,very thick,fit= (s_2_dummy_a)(s_3_dummy_a)  (s_2)] {};
\end{scope}

\begin{scope}[on background layer]
\node [xshift=6mm,yshift=6mm,model,red!10,fill=red!1,very thick,fit=(tok-0) (o-4)  ]{};
\end{scope}

\begin{scope}[on background layer]
\node [xshift=4mm,yshift=4mm,model,red!23,fill=red!2.3,very thick,fit=(tok-0) (o-4) ]{};
\end{scope}

\begin{scope}[on background layer]
\node [xshift=2mm,yshift=2mm,model,red!36,fill=red!3.6,very thick,fit=(tok-0) (o-4) ]{};
\end{scope}

\begin{scope}[on background layer]
\node [model,red!50,fill=red!5,very thick,fit=(tok-0) (o-4)]{};
\end{scope}

\begin{scope}[on background layer]
\node [model,blue!50,fill=blue!5,very thick,fit=(s_3_dummy) (s_4_dummy) (bert)]{};
\end{scope}

\begin{scope}[on background layer]
\node ()[sum,dotted,thick,fit=(o-0)(tok-0)] {};
\end{scope}

\begin{scope}[on background layer]
\node ()[sum,dotted,thick,fit=(o-2)(tok-2)] {};
\end{scope}

\begin{scope}[on background layer]
\node ()[sum,dotted,thick,fit=(o-4)(tok-4)] {};
\end{scope}

\node (p1) [below right=0.01mm and 0.01mm of first,xshift=-2mm,yshift=-2mm] {};
\node (p2) [below right=0.01mm and 0.01mm of last,xshift=2mm,yshift=2mm] {};
\draw[draw,shorten >= -1mm,path fading=east] (p1) to (p2) ;
	\foreach \x / \color in {0.2/black!85,0.4/black!60,0.6/black!40,0.8/black!15}
\draw [color=\color]($(p1)!\x!(p2)$) ++ (-2pt,+ 2pt) to + (+4pt, -4pt);
\node (t_j) [below right=0.01mm and 0.01mm of first,xshift=0.5mm,yshift=-1.5mm] {$t_j$};

\end{tikzpicture}
        \vspace{-0.2cm}
        \caption{Model architecture. Words are mapped to dynamic embeddings by the parts of the dynamic component (\tikz[baseline=-0.6ex] \node[rectangle,minimum width=0.5cm,inner sep=4pt,red!50,very thick,draw,rounded corners=3pt,fill=red!5,very thick] (0, 3) {};), which are then contextualized
        by the contextualizer (\tikz[baseline=-0.6ex] \node[rectangle,minimum width=0.5cm,blue!50,very thick,draw,inner sep=4pt,rounded corners=3pt,fill=blue!5,very thick] (0, 3) {};). The output of the contextualizer is used to compute the task-specific loss $\mathcal{L}_{\task}$.} \label{fig:architecture}
\end{figure}

Since many words in the core vocabulary are semantically stable
across social and temporal contexts, we
place a Gaussian prior on 
$\mathbf{e}^{(k)}_{ij}$,
\begin{equation} \label{eq:dist-e_k}
\mathbf{e}^{(k)}_{ij} \sim \mathcal{N} \left( \tilde{ \mathbf{e} }^{(k)}, \lambda_a^{-1} \mathbf{I} \right),
\end{equation}
where $\tilde{ \mathbf{e} }^{(k)}$ denotes a non-dynamic representation of
$x^{(k)}$.
Combining Equations \ref{eq:d} and \ref{eq:dist-e_k}, we write the function $d$ as
\begin{equation} \label{eq:e-o-sum}
d \left( x^{(k)}, s_i, t_j  \right) = \tilde{ \mathbf{e} }^{(k)} + \mathbf{o}^{(k)}_{ij},
\end{equation}
where $\mathbf{o}^{(k)}_{ij}$ denotes the vector offset from
$x^{(k)}$'s non-dynamic embedding $\tilde{ \mathbf{e} }^{(k)}$, which is 
stable across social and temporal contexts,
to its
dynamic embedding  $\mathbf{e}^{(k)}_{ij}$, which is specific to $s_i$ and $t_j$. The distribution of 
$\mathbf{o}^{(k)}_{ij}$ then follows
a Gaussian with
\begin{equation} \label{eq:o-prior}
\mathbf{o}^{(k)}_{ij} \sim \mathcal{N} \left(\mathbf{0}, \lambda_a^{-1} \mathbf{I} \right).
\end{equation}
We enforce Equation \ref{eq:o-prior} by including 
a regularization term in the objective function (Section \ref{sec:training}).

\subsection{Contextualizing Component} \label{sec:contextualizer}

We leverage a PLM for the function $c$,
specifically BERT \citep{Devlin.2019}. Denoting with $E_{ij}$ the sequence of dynamic embeddings corresponding to $X$ in $s_i$ and $t_j$,
the dynamic version of
Equation~\ref{eq:context} becomes
\begin{equation}
\mathbf{h}^{(k)}_{ij} = \BERT \left( \mathbf{e}^{(k)}_{ij}, E_{ij}^{(< k)}, E_{ij}^{(> k)} \right).
\end{equation}
We also use BERT, specifically its pretrained
input embeddings, to 
initialize the non-dynamic 
embeddings $\tilde{ \mathbf{e}}^{(k)}$, which are summed with 
the vector offsets $\mathbf{o}^{(k)}_{ij}$ (Equation 
\ref{eq:e-o-sum})
and fed into BERT.

Using a PLM for $c$ has the advantage
of making it easy to employ dynamic contextualized
word embeddings for downstream tasks by 
adding a task-specific layer on top 
of the PLM.

\subsection{Dynamic Component}

We model the vector offset $\mathbf{o}^{(k)}_{ij}$ 
as a function
of the word $x^{(k)}$, which we represent 
by its non-dynamic embedding $\tilde{ \mathbf{e}}^{(k)}$, as well as the social context $s_i$, 
which we represent by a time-specific embedding $\mathbf{s}_{ij}$. We use
BERT's pretrained input embeddings for $\tilde{ \mathbf{e}}^{(k)}$.\footnote{We also 
tried to learn separate embeddings in the dynamic component,
but this
led to worse performance.} We combine these representations in a time-specific feed-forward network,
\begin{equation}
\mathbf{o}^{(k)}_{ij} = \FFN_j \left( \tilde{ \mathbf{e} }^{(k)}  \!
\concat \mathbf{s}_{ij} \right),
\end{equation}
where $\concat$ denotes concatenation. 
To compute the social embedding $\mathbf{s}_{ij}$, 
we follow common practice in the computational
social sciences and represent the social 
community as a graph $\mathcal{G} = (\mathcal{S}, \mathcal{E})$,
where $\mathcal{S}$ is the set of social units $s_i$, and $\mathcal{E}$ is the 
set of edges between them (Section \ref{sec:data}).
We use a time-specific graph attention network (GAT) as proposed by \citet{Velickovic.2018}
to encode $\mathcal{G}$,\footnote{We also tried a model with a feed-forward network instead of graph attention, 
but it consistently performed worse.} 
\begin{equation}
\mathbf{s}_{ij} = \GAT_j \left( \tilde{\mathbf{s}}_i, \mathcal{G} \right).
\end{equation}
We initialize $\tilde{\mathbf{s}}_i$ with 
node2vec \citep{Grover.2016} embeddings.

To model the temporal drift of the dynamic embeddings $\mathbf{e}^{(k)}_{ij}$, we follow previous work on dynamic word embeddings \citep{Bamler.2017, Rudolph.2018} and impose a random walk prior over $\mathbf{o}^{(k)}_{ij}$,
\begin{equation}
\mathbf{o}^{(k)}_{ij} \sim \mathcal{N} \left(\mathbf{o}^{(k)}_{ij'}, \lambda_w^{-1} \mathbf{I} \right),
\end{equation}
with $j' = j - 1$. This type of Gaussian process is known as Ornstein-Uhlenbeck process \citep{Uhlenbeck.1930} and is commonly used to model time series \citep{Roberts.2013}. The random walk prior enforces that the dynamic embeddings $\mathbf{e}^{(k)}_{ij}$ change smoothly over time.

\begin{table*} [t!]\centering
\resizebox{\linewidth}{!}{%
\begin{tabular}{@{}lrlrlrrrrrlrrr@{}}
\toprule
{} & {} & \multicolumn{2}{c}{Linguistic} & \multicolumn{6}{c}{Social}  &  \multicolumn{4}{c}{Temporal} \\
\cmidrule(lr){3-4}
\cmidrule(lr){5-10}
\cmidrule(l){11-14}
Dataset & $|\mathcal{D}|$ & Unit & $\mu_{|X|}$& Unit & $|\mathcal{S}|$ & $|\mathcal{E}|$  & $\mu_d$  & $\mu_\pi$ & $\rho$ & Unit & $|\mathcal{T}|$ & $t_1$ & $t_{|\mathcal{T}|}$ \\
\midrule
ArXiv & 972,369 & Abstract& 118.10 & Subject & 535 & 5,165 & 19.34 & 3.48 & .036 & Year  & 20 & [01/]2001 & [10/]2020 \\
Ciao & 269,807 &Review &684.68 & User & 10,880 & 129,900 & 18.20 &  3.65 & .002 & Year & 12 & [05/]2000 & [09/]2011 \\
Reddit & 915,663 & Comment & 43.50 & Subreddit & 5,728 & 61,796& 23.99 &4.69 & .005 & Month  & 8 & 09/2019 & 04/2020\\
YELP & 795,661 & Review&151.59 &  User & 5,203 & 223,254 & 45.17 & 2.83& .009 & Year   & 10 & [01/]2010 & [12/]2019 \\
\bottomrule
\end{tabular}}
\caption{Dataset statistics. $|\mathcal{D}|$: number of data points; $\mu_{|X|}$: average number of tokens per text; $|\mathcal{S}|$: number of nodes in network; $|\mathcal{E}|$: number of 
edges; $\mu_d$: average node degree; $\mu_\pi$: average shortest path length between two nodes; $\rho$: network density; $|\mathcal{T}|$: number of time points; $t_1$: first time point; $t_{|\mathcal{T}|}$: last time point. In cases where years are the temporal unit, we also provide the first and last month included in the data.}  \label{tab:data-stats}
\end{table*}

\subsection{Model Training} \label{sec:training}

The combination with BERT makes dynamic contextualized word embeddings
easily applicable to different tasks by adding a task-specific
layer on top 
of the contextualizing component.
For training the model, the overall
loss is
\begin{equation}
\mathcal{L}_{\total} = \mathcal{L}_{\task} + \mathcal{L}_{\prior_a} +  \mathcal{L}_{\prior_w} ,
\end{equation}
where $\mathcal{L}_{\task}$ is the task-specific loss, and $\mathcal{L}_{\prior_a}$ and $\mathcal{L}_{\prior_w}$ are the regularization terms that impose the anchoring and random walk priors on the
type-level
offset vectors,
\begin{align}
\mathcal{L}_{\prior_a} & =   \frac{\lambda_a}{K}\sum_{k=1}^K  \mathbin{\|} \mathbf{o}^{(k)}_{ij}  \! \mathbin{\|} ^2_2 \\  \mathcal{L}_{\prior_w} & = \frac{\lambda_w}{K}\sum_{k=1}^K   \mathbin{\|} \mathbf{o}^{(k)}_{ij} - \mathbf{o}^{(k)}_{ij'} \! \mathbin{\|} ^2_2.
\end{align}
It is common practice to set $\lambda_a \ll \lambda_w$
\citep{Bamler.2017, Rudolph.2018}. Here, we set 
$\lambda_a = 10 ^{-3} \cdot  \lambda_w$, which reduces the number of tunable hyperparameters.
We place
the priors only on frequent words in
the vocabulary (Section \ref{sec:embedding-training}), taking into account the observation that the vocabulary core
constitutes the best basis
for dynamic word embeddings \citep{Hamilton.2016}.

\section{Data} \label{sec:data}

We fit dynamic contextualized word embeddings to four datasets with different linguistic,
social, and temporal characteristics, which allows us to investigate
factors impacting their utility. Each dataset $\mathcal{D}$
consists of a set of texts (e.g., reviews) written by a set of 
social units $\mathcal{S}$ (e.g., users) over a sequence of time periods $\mathcal{T}$ (e.g., years).
Furthermore, the social units are connected by a set of edges $\mathcal{E}$ within a social network $\mathcal{G}$.
Table \ref{tab:data-stats} provides summary statistics of the four datasets.

\textbf{ArXiv.} ArXiv is an open-access distribution service for scientific 
articles. Recently, a dataset of all papers published on ArXiv with corresponding metadata was released.\footnote{\url{https://www.kaggle.com/Cornell-University/arxiv}} 
For this study, we use ArXiv's subject classes (e.g., \texttt{cs.CL}) as social units 
and extract the abstracts of papers published between 2001 and 2020 for subjects with
at least 100 publications in that time.\footnote{We treat subject class combinations passing
the frequency threshold
(e.g., \texttt{cs.CL\&cs.AI}) as individual units.} 
To create the network, we measure the overlap in authors between subject classes as
the Jaccard similarity of corresponding author sets, resulting in a similarity matrix $\mathbf{S}$. Based on $\mathbf{S}$, we define the adjacency matrix $\mathbf{G}$ of $\mathcal{G}$, whose elements are
\begin{equation}
G_{ij} =  \bigl\lceil S_{ij} - \theta \bigr\rceil,
\end{equation}
i.e., there is an edge between subject classes $i$ and $j$ if the Jaccard similarity of author sets is greater than $\theta$. We set $\theta$ to 0.01.\footnote{We tried other values of $\theta$, but the results were similar.}

\textbf{Ciao.} Ciao is a product review site on which users can mark explicit trust relations towards 
other users (e.g., if they 
 find their reviews helpful). A 
dataset containing reviews covering the time period from 2000 to 2011 has been made publicly available \citep{Tang.2012}.\footnote{\url{https://www.cse.msu.edu/~tangjili/trust.html}}
We use the trust relations to create a directed graph. Since we also perform sentiment analysis on the dataset, we follow \citet{Yang.2017} in converting the five-star rating range into two classes by discarding three-star reviews and treating four/five stars as positive and one/two stars as negative.

\textbf{Reddit.} Reddit is a social media
platform hosting discussions about a variety of topics. It is divided into smaller communities, so-called subreddits, which have been shown to be highly conducive to linguistic dynamics \citep{DelTredici.2018, DelTredici.2019}. A full dump of public Reddit posts is available online.\footnote{\url{https://files.pushshift.io/reddit/comments}} We retrieve 
all comments between September 2019 and April 2020, which 
allows us to 
examine the effects of the rising Covid-19 pandemic on lexical usage patterns. 
We remove subreddits with fewer than 10,000 comments in the examined time period and sample 20 comments per subreddit and month.
For each subreddit, we compute the set of users with at least 10 comments in the examined time period. 
Based on this, we use the same strategy as for ArXiv to create a network based on user overlap.

\textbf{YELP.} Similarly to Ciao, YELP is a product review site on which users can mark explicit friendship relations. A subset of the data has been released online.\footnote{\url{https://www.yelp.com/dataset}} We use the friendship relations to create a directed graph between users. Since we also use the dataset for sentiment analysis, we again discard three-star reviews and convert the five-star rating range into two classes.

The fact that the datasets differ 
in terms of their social and temporal characteristics allows us to examine which factors
impact
the utility of dynamic contextualized word 
embeddings. We highlight, e.g., that the datasets differ in the nature of their social units, cover different time periods, and exhibit different levels of temporal granularity.
We randomly split all datasets into 70\% training, 10\%
development, and 20\% test. We apply stratified sampling 
to make sure the model sees data from all time points 
during training. See Appendix \ref{app:preprocessing} for details about data preprocessing.

\section{Experiments}

\subsection{Embedding Training} \label{sec:embedding-training}

\begin{table} [t!]\centering
\resizebox{\linewidth}{!}{%
\begin{tabular}{@{}lrrrrrrrr@{}}
\toprule
{} & \multicolumn{2}{c}{ArXiv}  & \multicolumn{2}{c}{ Ciao} & \multicolumn{2}{c}{Reddit}  & \multicolumn{2}{c}{ YELP}\\
\cmidrule(lr){2-3}
\cmidrule(lr){4-5}
\cmidrule(lr){6-7}
\cmidrule(l){8-9}
Model & Dev & Test & Dev & Test  & Dev & Test & Dev & Test \\
\midrule
DCWE & \best{3.521} &  \best{{\underline{3.513}}} & \best{5.920}  & \best{5.902} &  \best{{\underline{9.480}}}  & 9.596 &  4.717 & \best{4.720}  \\
CWE &  3.523  & 3.530  &   5.922 & 5.910 &  9.580 & \best{9.555} & \best{4.714} & 4.723   \\ 
\bottomrule
\end{tabular}
}
\caption{Masked language modeling perplexity on the four datasets (lower is better). DCWE:
dynamic contextualized word embeddings; CWE: contextualized word embeddings. 
The better score per column (highlighted in gray)
is underlined if it is significantly ($p < .01$) better as shown by a Wilcoxon signed-rank test.}  \label{tab:performance}
\end{table}

We fit dynamic contextualized word embeddings to all four datasets, using 
BERT\textsubscript{BASE} (uncased) as the contextualizer and
masked language modeling as the training objective \citep{Devlin.2019}, i.e., 
we add a language modeling head on top of BERT.\footnote{For
  a given dataset, we only compute dynamic embeddings for tokens in BERT's input vocabulary 
that are among the 100,000 most frequent words. For less frequent tokens, 
we input the non-dynamic BERT embedding.}
To estimate the goodness of fit, we measure masked language modeling perplexity and compare against 
finetuned (non-dynamic) contextualized word embeddings, specifically BERT\textsubscript{BASE} (uncased).
See Appendix \ref{app:hyper} for details 
about implementation, hyperparameter tuning, and runtime.

Dynamic contextualized word embeddings (DCWE) yield fits to the data similar to
and (sometimes significantly) better than non-dynamic contextualized word
embeddings (CWE), which indicates that they successfully combine
extralinguistic with linguistic information (Table \ref{tab:performance}).\footnote{Statistical significance is tested with a 
Wilcoxon signed-rank test \citep{Wilcoxon.1945, Dror.2018}.}

\subsection{Ablation Study}

\begin{figure}
        \centering
        \begin{subfigure}[b]{0.225\textwidth}
            \centering
            \begin{tikzpicture}
[xscale=0.9,
 thick,
 model/.style={draw,rectangle,inner sep=6pt,rounded corners=6pt},
 sum/.style={draw,rectangle,inner sep=2pt,rounded corners=6pt},
 vec/.style={draw,rectangle,fill=white,inner sep=0pt,minimum width=1.5cm,minimum height=0.6cm,text height=0.3cm,text depth=0.05cm,rounded corners=8pt},
  token/.style={draw,rectangle,fill=white,inner sep=0pt,minimum width=0.65cm,minimum height=0.6cm,text height=0.3cm,text depth=0.05cm},
  comp/.style={draw,rectangle,fill=white,inner sep=0pt,minimum width=5cm,minimum height=1cm,text height=0.3cm,text depth=0.05cm},
 place/.style={circle,draw,fill=white,
inner sep=0pt,minimum size=5mm}]

\pgfmathsetseed{1} 

\foreach \x / \xtext in {0/$\tilde{\mathbf{e}}^{(1)}$,2/$\tilde{\mathbf{e}}^{(2)}$,4/$\tilde{\mathbf{e}}^{(3)}$}
	\node [vec] (tok-\x) at (\x + 0.375,1) {\xtext};
	
\foreach \x in {0,2,4}
	\node [] () at (\x + 0.375,0.5) {$+$};
	
\foreach \x / \xtext in {0/$\mathbf{o}_{j}^{(1)}$,2/$\mathbf{o}_{j}^{(2)}$,4/$\mathbf{o}_{j}^{(3)}$}
	\node [vec] (o-\x) at (\x + 0.375,0) {\xtext};

	\foreach \pos / \x / \xtext in {-.1/0/$x^{(1)}$,0.9/1/$x^{(2)}$,1.9/2/$x^{(3)}$}
	\node [token] (x-\x) at (\pos ,-1.5) {\xtext};

\node[comp](bert)[above=0.7cm of tok-2, xshift=3mm,yshift=3mm] {$\BERT$};

\node[](loss)[above=0.7cm of bert] {$\mathcal{L}_{\task}$};

\node[vec,draw=none,fill=none](s_2_dummy_a)[below=0.7cm of o-4] {};
\node[vec,draw=none,fill=none](s_3_dummy_a)[below=1cm of o-4] {};

\node[token,draw=none,fill=none](s_1_dummy_b)[below=0.7cm of o-2] {};
\node[vec,draw=none,fill=none](s_2_dummy_b)[below=0.7cm of o-0] {};
\node[vec,draw=none,fill=none](s_3_dummy_b)[below=1cm of o-0] {};

\node[vec,draw=none,fill=none](s_3_dummy)[above=0.7cm of tok-0,xshift=3mm,yshift=3mm] {};
\node[vec,draw=none,fill=none](s_4_dummy)[above=0.7cm of tok-4,xshift=3mm,yshift=3mm] {};

	\path (x-0) edge[->,in=180,out=180,dashed] ([xshift=-.257cm]o-0.west);
		\path (x-0) edge[->,in=180,out=180,dashed] ([xshift=-.257cm]tok-0.west);
		
	\path (bert) edge[->,dashed] (loss);
		
\path ([yshift=.075cm]tok-0.north) edge[->,in=-90,out=90,dashed] ([xshift=-1.2cm,yshift=-.234cm]bert.south);

\begin{scope}[on background layer]
\node (last)[xshift=6mm,yshift=6mm,model,red!10,fill=red!1,very thick,fit=(x-2)(s_2_dummy_b) (s_3_dummy_b)] {};
\end{scope}

\begin{scope}[on background layer]
\node ()[xshift=4mm,yshift=4mm,model,red!23,fill=red!2.3,very thick,fit=(x-2)(s_2_dummy_b) (s_3_dummy_b)] {};
\end{scope}

\begin{scope}[on background layer]
\node ()[xshift=2mm,yshift=2mm,model,red!36,fill=red!3.6,very thick,fit=(x-2)(s_2_dummy_b) (s_3_dummy_b)] {};
\end{scope}

\begin{scope}[on background layer]
\node (first)[model,red!50,fill=red!5,very thick,fit=(x-2)(s_2_dummy_b) (s_3_dummy_b)] {};
\end{scope}

\begin{scope}[on background layer]
\node [xshift=6mm,yshift=6mm,model,red!10,fill=red!1,very thick,fit=(tok-0) (o-4)  ]{};
\end{scope}

\begin{scope}[on background layer]
\node [xshift=4mm,yshift=4mm,model,red!23,fill=red!2.3,very thick,fit=(tok-0) (o-4) ]{};
\end{scope}

\begin{scope}[on background layer]
\node [xshift=2mm,yshift=2mm,model,red!36,fill=red!3.6,very thick,fit=(tok-0) (o-4) ]{};
\end{scope}

\begin{scope}[on background layer]
\node [model,red!50,fill=red!5,very thick,fit=(tok-0) (o-4)]{};
\end{scope}

\begin{scope}[on background layer]
\node [model,blue!50,fill=blue!5,very thick,fit=(s_3_dummy) (s_4_dummy) (bert)]{};
\end{scope}

\begin{scope}[on background layer]
\node ()[sum,dotted,thick,fit=(o-0)(tok-0)] {};
\end{scope}

\begin{scope}[on background layer]
\node ()[sum,dotted,thick,fit=(o-2)(tok-2)] {};
\end{scope}

\begin{scope}[on background layer]
\node ()[sum,dotted,thick,fit=(o-4)(tok-4)] {};
\end{scope}

\node (p1) [below right=0.01mm and 0.01mm of first,xshift=-2mm,yshift=-2mm] {};
\node (p2) [below right=0.01mm and 0.01mm of last,xshift=2mm,yshift=2mm] {};
\draw[draw,shorten >= -1mm,path fading=east] (p1) to (p2) ;
	\foreach \x / \color in {0.2/black!85,0.4/black!60,0.6/black!40,0.8/black!15}
\draw [color=\color]($(p1)!\x!(p2)$) ++ (-2pt,+ 2pt) to + (+4pt, -4pt);
\node (t_j) [below right=0.01mm and 0.01mm of first,xshift=0.5mm,yshift=-1.5mm] {$t_j$};

\end{tikzpicture}
  \caption[]%
            {{\small Social ablation}}   
  \label{fig:s-abl}
        \end{subfigure}
        \begin{subfigure}[b]{0.225\textwidth}  
            \centering 
            \begin{tikzpicture}
[xscale=0.9,
 thick,
 model/.style={draw,rectangle,inner sep=6pt,rounded corners=6pt},
 sum/.style={draw,rectangle,inner sep=2pt,rounded corners=6pt},
 vec/.style={draw,rectangle,fill=white,inner sep=0pt,minimum width=1.5cm,minimum height=0.6cm,text height=0.3cm,text depth=0.05cm,rounded corners=8pt},
  token/.style={draw,rectangle,fill=white,inner sep=0pt,minimum width=0.65cm,minimum height=0.6cm,text height=0.3cm,text depth=0.05cm},
  comp/.style={draw,rectangle,fill=white,inner sep=0pt,minimum width=5cm,minimum height=1cm,text height=0.3cm,text depth=0.05cm},
 place/.style={circle,draw,fill=white,
inner sep=0pt,minimum size=5mm}]

\pgfmathsetseed{1} 

\foreach \x / \xtext in {0/$\tilde{\mathbf{e}}^{(1)}$,2/$\tilde{\mathbf{e}}^{(2)}$,4/$\tilde{\mathbf{e}}^{(3)}$}
	\node [vec] (tok-\x) at (\x + 0.375,1) {\xtext};
	
\foreach \x in {0,2,4}
	\node [] () at (\x + 0.375,0.5) {$+$};
	
\foreach \x / \xtext in {0/$\mathbf{o}_{i}^{(1)}$,2/$\mathbf{o}_{i}^{(2)}$,4/$\mathbf{o}_{i}^{(3)}$}
	\node [vec] (o-\x) at (\x + 0.375,0) {\xtext};

	\foreach \pos / \x / \xtext in {-.1/0/$x^{(1)}$,0.9/1/$x^{(2)}$,1.9/2/$x^{(3)}$}
	\node [token] (x-\x) at (\pos ,-1.5) {\xtext};

\node[comp](bert)[above=0.7cm of tok-2] {$\BERT$};

\node[](loss)[above=0.7cm of bert] {$\mathcal{L}_{\task}$};

\node[vec,draw=none,fill=none](s_2_dummy_a)[below=0.7cm of o-4] {};
\node[vec,draw=none,fill=none](s_3_dummy_a)[below=1cm of o-4] {};

\node[token,draw=none,fill=none](s_1_dummy_b)[below=0.7cm of o-2] {};
\node[vec,draw=none,fill=none](s_2_dummy_b)[below=0.7cm of o-0] {};
\node[vec,draw=none,fill=none](s_3_dummy_b)[below=1cm of o-0] {};

\node[vec,draw=none,fill=none](s_3_dummy)[above=0.7cm of tok-0] {};
\node[vec,draw=none,fill=none](s_4_dummy)[above=0.7cm of tok-4] {};

\node[place](s_1)[below=0.65cm of o-4] {$s_i$};
\node[place](s_2)[left=0.42cm of s_1,yshift=-0.2cm] {};
\node[place](s_3)[below left=0.15cm and 0.05cm of s_1]  {};
\node[place](s_4) [below right=0.1cm and 0.12cm of s_1]  {};

	\path (s_1) edge[->,in=-90,out=90,dashed] ([yshift=-.234cm]o-0.south);
	\path (x-0) edge[->,in=180,out=180,dashed] ([xshift=-.257cm]o-0.west);
		\path (x-0) edge[->,in=180,out=180,dashed] ([xshift=-.257cm]tok-0.west);
		
	\path (bert) edge[->,dashed] (loss);
		
\path ([yshift=.075cm]tok-0.north) edge[->,in=-90,out=90,dashed] ([xshift=-1.2cm,yshift=-.234cm]bert.south);

\draw[](s_1) to (s_2);
\draw[](s_1) to (s_3);
\draw[](s_1) to (s_4);

\begin{scope}[on background layer]
\node ()[model,red!50,fill=red!5,very thick,fit=(x-2)(s_2_dummy_b) (s_3_dummy_b)] {};
\end{scope}

\begin{scope}[on background layer]
\node (first)[model,red!50,fill=red!5,very thick,fit= (s_2_dummy_a)(s_3_dummy_a)  (s_2)] {};
\end{scope}

\begin{scope}[on background layer]
\node [model,red!50,fill=red!5,very thick,fit=(tok-0) (o-4)]{};
\end{scope}

\begin{scope}[on background layer]
\node [model,blue!50,fill=blue!5,very thick,fit=(s_3_dummy) (s_4_dummy) (bert)]{};
\end{scope}

\begin{scope}[on background layer]
\node ()[sum,dotted,thick,fit=(o-0)(tok-0)] {};
\end{scope}

\begin{scope}[on background layer]
\node ()[sum,dotted,thick,fit=(o-2)(tok-2)] {};
\end{scope}

\begin{scope}[on background layer]
\node ()[sum,dotted,thick,fit=(o-4)(tok-4)] {};
\end{scope}

\end{tikzpicture}
            \vspace{0.66cm}
            \caption[]%
            {{\small Temporal ablation}}    
            \label{fig:t-abl}
        \end{subfigure}
      
        \caption[]{Models for ablation study. In social ablation, the vector offset only depends on word identity and time, not social context. In temporal ablation, there is only one social component
        for all time slices.}
        \label{fig:abl}
\end{figure}
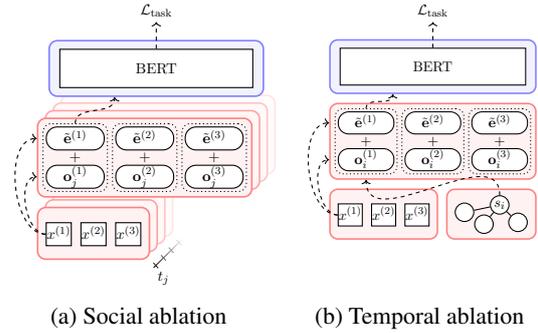

To examine the relative importance of temporal and social 
information for dynamic contextualized word embeddings,
we perform two experiments in which we ablate social context and time (Figure \ref{fig:abl}). In social ablation (SA),
we train dynamic contextualized word embeddings where
the vector offset depends only on word identity and
time, not social context, keeping the random walk 
prior between subsequent time slices. In temporal ablation (TA), we use one social component 
for all 
time slices. See Appendix \ref{app:hyper-abl} for details 
about implementation, hyperparameter tuning, and runtime.

Temporal ablation has more severe consequences than social ablation
(Table \ref{tab:ablation}). On Ciao, the social component does not yield better fits on the data at all, which might be related to the fact that many users in this dataset 
only have one
review, and that its social network has the lowest density as well as the 
smallest average node degree out of all considered datasets (Table \ref{tab:data-stats}).

\begin{table} [t!]\centering
\resizebox{\linewidth}{!}{%
\begin{tabular}{@{}lrrrrrrrr@{}}
\toprule
{} & \multicolumn{2}{c}{ArXiv}  & \multicolumn{2}{c}{ Ciao} & \multicolumn{2}{c}{Reddit}  & \multicolumn{2}{c}{ YELP}\\
\cmidrule(lr){2-3}
\cmidrule(lr){4-5}
\cmidrule(lr){6-7}
\cmidrule(l){8-9}
Model & Dev & Test & Dev & Test  & Dev & Test & Dev & Test \\
\midrule
DCWE & 3.521 & \best{3.513} & 5.920  & 5.902 &  \best{{\underline{9.480}}}  & \best{9.596} &  \best{{\underline{4.717}}} & \best{4.720}  \\
SA &  \best{3.517}  & 3.515 &  \best{5.919} &  \best{5.899} &     9.620  &  9.631 &  4.725 & 4.723 \\ 
TA &  3.534  &   3.541 &  5.924 & 5.931 &  9.598 & 9.612 & 4.726 &  4.734  \\ 

\bottomrule
\end{tabular}
}
\caption{Masked language modeling perplexity on the four datasets in ablation study (lower is better). DCWE: dynamic contextualized word embeddings; SA: social ablation; TA: temporal ablation. The best score per column (highlighted in gray)
is underlined if it is significantly ($p < .01$) better than the second-best 
score as shown by a Wilcoxon signed-rank test.}  \label{tab:ablation}
\end{table}

\subsection{Qualitative Analysis} \label{sec:qual-analyis}

\begin{table*} [t!]\centering
\resizebox{\linewidth}{!}{%
\begin{tabular}{@{}lllll@{}}
\toprule
{} &  \multicolumn{2}{c}{Context for $\cossim^{(k)}_{ij} > \mu^{(k)}_{\cossim} $\phantom{xxxxxxxxxx}}  &  \multicolumn{2}{c}{Context for $\cossim^{(k)}_{ij} < \mu^{(k)}_{\cossim} $\phantom{xxxxxxxxxxxxx}}  \\
\cmidrule(lr){2-3}
\cmidrule(lr){4-5}
Word & Extralinguistic & Linguistic & Extralinguistic & Linguistic \\
\midrule
``isolating'' & \makecell[l]{ \texttt{r/SAHP} \\ 12/19 } & \textit{ \makecell[l]{It's really hard to explain 
to other people\\how isolating and exhausting being a\\SAHP can be.}  } &
\makecell[l]{ \texttt{r/Asthma} \\ 03/20} & \textit{ \makecell[l]{ I wish I knew if I'd had covid so that
I\\could stop self isolating and instead\\volunteer in my community.}  } \\
\midrule
``testing'' & \makecell[l]{ \texttt{r/VJoeShows} \\ 04/20} & \textit{ \makecell[l]{Testing a photocell 
light fixture during\\the day is easy when you know how.\\This is what this DIY video is about.}  } &
\makecell[l]{ \texttt{r/vancouver} \\ 03/20 } & \textit{ \makecell[l]{ Testing is not required if a patient has 
no\\symptoms, mild symptoms, or is a returning\\traveller and is isolating at home.}  } \\
\bottomrule
\end{tabular}}
\caption{Examples of dynamics in word meaning during the Covid-19 pandemic.
The table lists example words with top-ranked values of $\sigma ^{(k)}_{\cossim}$, 
i.e., they exhibit a high degree of extralinguistically-driven semantic dynamics. }  \label{tab:dynamics-examples}
\end{table*}

Do dynamic contextualized word embeddings indeed capture 
interpretable dynamics in word meaning?
To examine this question qualitatively, we define as $\cossim^{(k)}_{ij}$
the cosine similarity between the non-dynamic embedding of $x^{(k)}$, 
$\tilde{\mathbf{e}}^{(k)}$,
and the dynamic embeddings of $x^{(k)}$ given 
social and temporal contexts $s_i$ and $t_j$, $\mathbf{e}^{(k)}_{ij}$,
\begin{equation}
\cossim^{(k)}_{ij} = \cos \phi^{(k)}_{ij},
\end{equation}
where $\phi_{ij}^{(k)}$ is the angle between $\tilde{\mathbf{e}}^{(k)}$ and $\mathbf{e}^{(k)}_{ij}$ (Figure \ref{fig:framework}).\footnote{In cases where $x^{(k)}$ is split into several WordPiece 
tokens by BERT, we follow previous work \citep{Pinter.2020b,
Sia.2020} and average the subword embeddings.} To find words with a high degree of variability,
we compute the standard deviation of $\cossim^{(k)}_{ij}$ based on all 
$s_i$ and $t_j$ in which a given word $x^{(k)}$ occurs in the data,
\begin{equation}
    \sigma ^{(k)}_{\cossim} = \sigma \left( \{  \cossim^{(k)}_{ij}  |  (x^{(k)}, s_i, t_j)  \in \mathcal{D}  \} \right),
\end{equation}
where we take the development set for $\mathcal{D}$.
Looking at the top-ranked 
words according to $\sigma ^{(k)}_{\cossim}$, we observe that
they 
exhibit pronounced extralinguistically-driven semantic dynamics in the data.
For Reddit, e.g., many of the top-ranked words
have experienced a sudden shift in their dominant sense 
during the Covid-19
pandemic such as ``isolating'' and ``testing'' (Table \ref{tab:dynamics-examples}).
Social and temporal contexts in which
the sense related to Covid-19 is dominant have
smaller values of  $\cossim^{(k)}_{ij}$ (i.e., the cosine distance is larger) than the ones
in which the more general sense is dominant. Such
short-term semantic shifts, which have
attracted growing interest in NLP recently \citep{Stewart.2017, DelTredici.2019, Powell.2020},
  can result 
in lasting semantic narrowing if speakers become reluctant
to use the word outside of the more specialized sense 
\citep{Anttila.1989, Croft.2000, Robinson.2012, Bybee.2015}.
 
Thus, the qualitative analysis suggests that the dynamic component
indeed captures
extralinguistically-driven variability in word meaning.
In Sections \ref{sec:exploration-1} and \ref{sec:exploration-2}, we will demonstrate
by means of two example applications how this property can be beneficial in practice.

\subsection{Exploration 1: Semantic Diffusion} \label{sec:exploration-1}

We will now provide a more in-depth analysis of social and temporal dynamics in word meaning to showcase the potential
of dynamic contextualized word embeddings as an analytical tool. Specifically, we will analyze how
changes in the dominant sense of a word diffuse through the social networks of ArXiv and Reddit. For ArXiv, we will examine the deep learning sense of the word ``network''. For Reddit, we will focus on the   medical sense of the word ``mask''. We know that these senses have become more widespread over the last few years (ArXiv) and months (Reddit), but we want to test if dynamic contextualized word embeddings can capture
this spread, and if they allow us to gain new insights about the spread of semantic associations through 
social networks in general.

To perform this analysis, let $r_{ij}^{(k, k')}$ be the rank of
$x^{(k')}$'s embedding among the $N$ nearest neighbors of $x^{(k)}$'s embedding, 
given social and temporal contexts $s_i$ and $t_j$. We then define as
\begin{equation}
\hat{r}_{ij}^{(k, k')} = N  - r_{ij}^{(k, k')} + 1
\end{equation}
a semantic similarity score between $x^{(k)}$ and $x^{(k')}$. $\hat{r}_{ij}^{(k, k')}$ is maximal 
when $x^{(k')}$'s embedding is closest to $x^{(k)}$'s embedding. We set $\hat{r}_{ij}^{(k, k')} = 0$ if $x^{(k')}$ is not among the $N$ nearest neighbors of $x^{(k)}$. 
We set $N = 100$.

\begin{figure*}
        \centering
        \begin{subfigure}[h]{\textwidth}
            \centering
        \includegraphics[width=\textwidth]{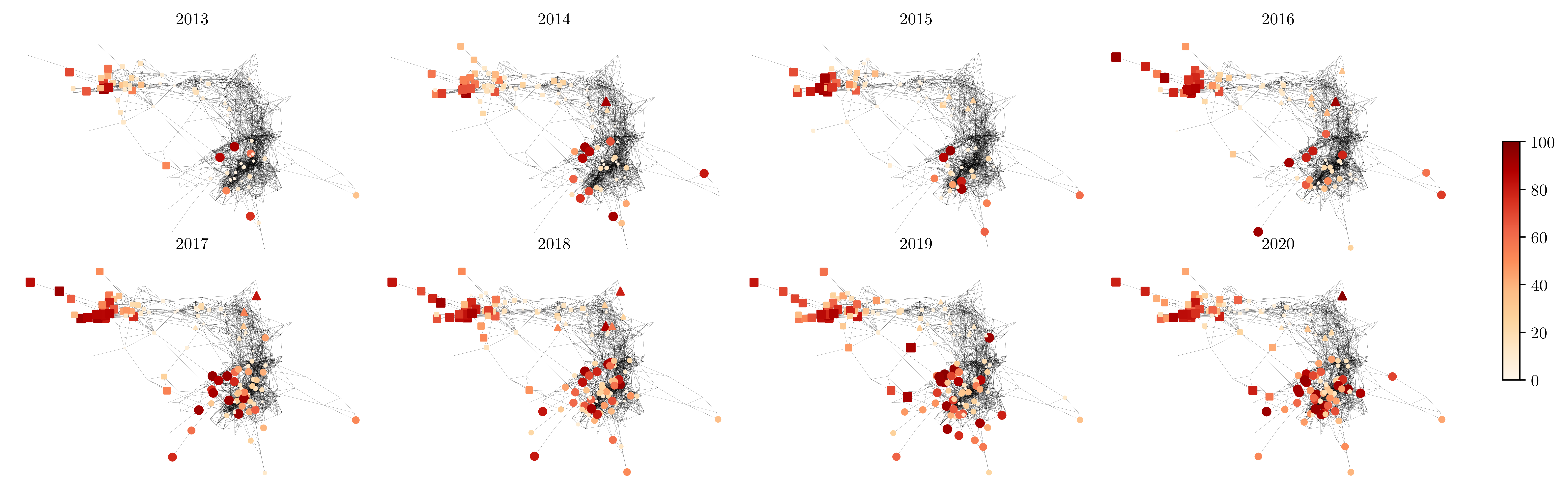}      
  \caption[]%
            {{\small $\hat{r}_{ij}^{(k, k')}$ for ``network'' and ``learning'' in ArXiv}}   
  \label{fig:spread-network}
                      \vspace{0.2cm}    
        \end{subfigure}
        \begin{subfigure}[h]{\textwidth}  
            \centering 
        \includegraphics[width=\textwidth]{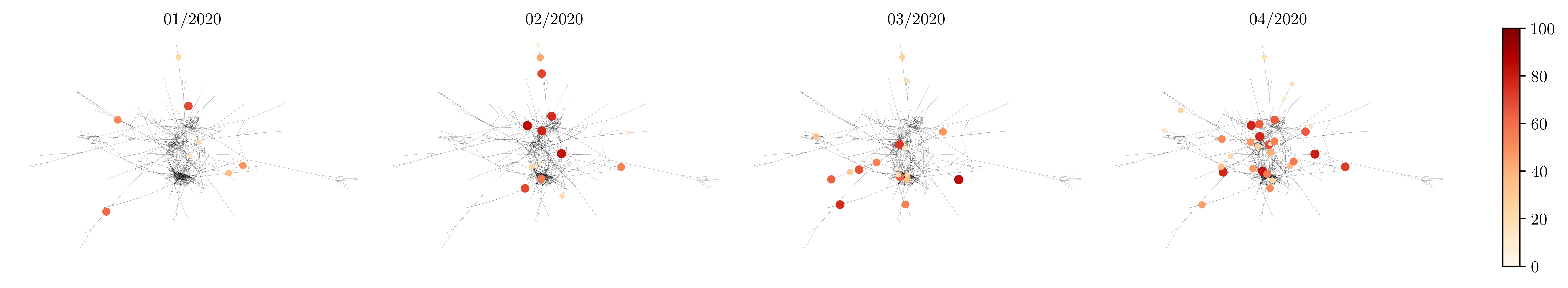}  
                    \vspace{-0.7cm}    
            \caption[]%
            {{\small $\hat{r}_{ij}^{(k, k')}$ for ``mask'' and ``vaccine'' in Reddit}}    
            \label{fig:spread-mask}
        \end{subfigure}
      
        \caption[]{Spread of changes in the dominant sense through the social network. The figure shows dynamics in $\hat{r}_{ij}^{(k, k')}$, a score for semantic similarity between 0 (no similarity) and 100 (very similar), for ``network'' and ``learning'' in ArXiv as well as ``mask'' and ``vaccine'' in Reddit. The different node shapes in the ArXiv network represent the three major ArXiv subject classes: computer science (square), mathematics (triangle), and physics (circle). For ``network'', the change towards the deep learning sense spread gradually from computer science and physics. For ``mask'', the change towards the medical sense also spread gradually, with a major intensification after 03/2020.}
        \label{fig:spread}
\end{figure*}

Using $\hat{r}_{ij}^{(k, k')}$, we measure dynamics in the semantic similarity between ``network'' and ``learning'' (representing the deep learning sense of ``network'') as well as ``mask'' and ``vaccine'' (representing the medical sense of ``mask''). 
For all social and temporal contexts in which ``network'' and ``mask''
occur, we compute $\hat{r}_{ij}^{(k, k')}$ between their socially and temporally dynamic embeddings on the one hand and 
time-specific centroids 
of ``learning'' and ``vaccine'' averaged over social contexts on the other, 
employing contextualized versions of the dynamic embeddings.\footnote{We average the first six layers of the contextualizer since they have been shown to contain the core of lexical and semantic information \citep{Vulic.2020}.} In cases where ``network'' or ``mask'' occur more than once in a certain social and temporal context, we take the mean of $\hat{r}_{ij}^{(k, k')}$.

The dynamics of $\hat{r}_{ij}^{(k, k')}$ 
reflect how the changes in the dominant sense of ``network'' and ``mask'' spread through the social networks (Figure \ref{fig:spread}). For ``network'', we see that the deep learning sense was already present in computer science and physics in 2013, where neural networks have been used since the 1980s. It then gradually spread from these two epicenters, with a major intensification after 2016. For ``mask'', we also see a gradual diffusion, with a major intensification after 03/2020.

\begin{figure}[t!]
        \centering
        \includegraphics[width=0.8\linewidth]{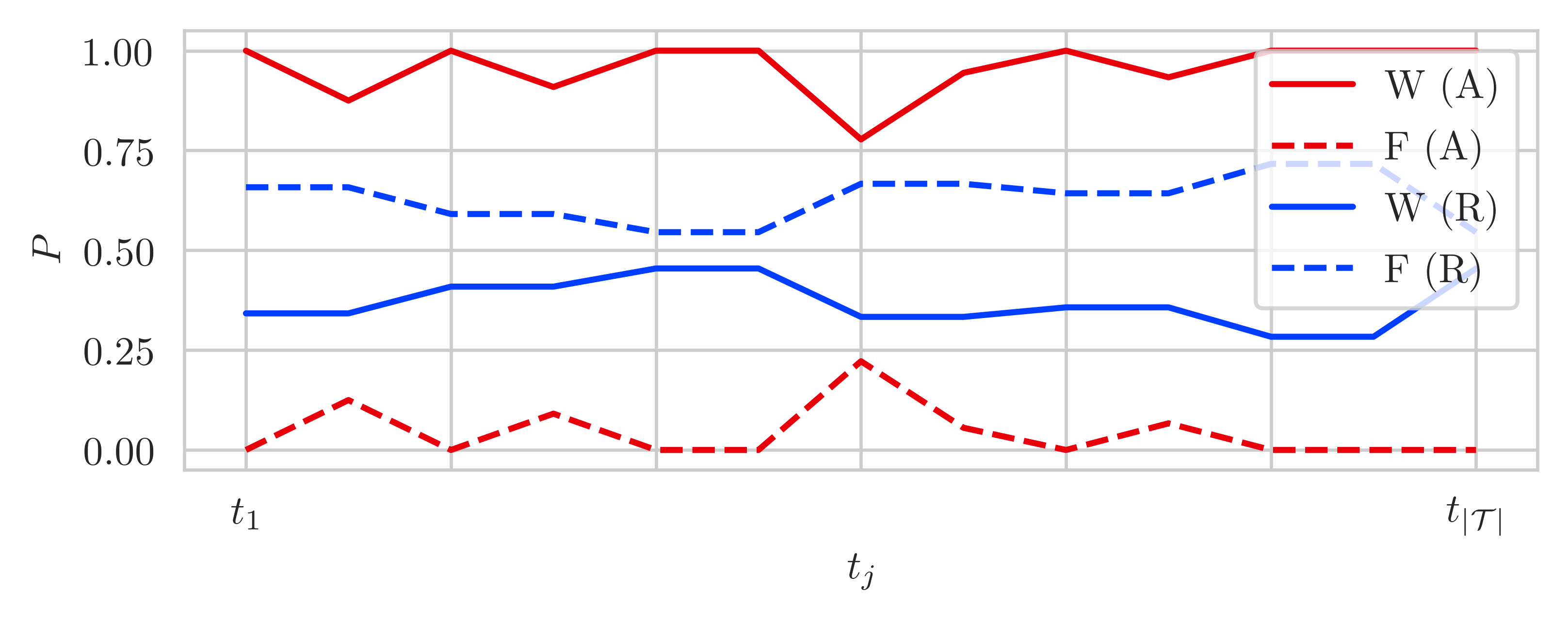} 
        \caption[]{Types of semantic diffusion in ArXiv (A) and Reddit (R). The figure shows for each time $t_j$ the probability that a node having the new sense for the first time is the neighbor of a node that already had it previously (walk, W) as opposed to cases where none of its neighbors had it previously (flight, F).}
        \label{fig:spread-probs}
\end{figure}

On what paths do new semantic associations spread through the social network? In complex systems theory, there
are two basic types of random motion on networks: random walks, which consist of a
series of consecutive random steps, and random flights, where step lengths are drawn from the Lévy distribution \citep{Masuda.2017}. To probe whether there is a dominant type of spread for the two examples, we compute for each time slice $t_j$ what proportion of nodes that have $\hat{r}_{ij}^{(k, k')} > 0$ for the first time at $t_j$ (i.e., the change in the dominant sense has just arrived) are neighbors of nodes that already had $\hat{r}_{ij}^{(k, k')} > 0$ before $t_j$. This analysis shows that random walks are the dominant type of spread for ``network'', but random flights for ``mask'' (Figure \ref{fig:spread-probs}). Intuitively, it makes sense that a technical concept such as neural networks spreads through the direct contact of collaborating scientists rather than through more distant forms of reception (e.g., the reading of articles). In the case of facial masks, on the other hand, the exogenous factor of the 
worsening Covid-19 pandemic and the accompanying publicity was a driver of semantic dynamics irrespective of node position.

\subsection{Exploration 2: Sentiment Analysis} \label{sec:exploration-2}

As a second testbed, we apply dynamic contextualized word embeddings on a task for which social and temporal information is known  
to be important \citep{Yang.2017}: sentiment analysis. We use the Ciao and YELP datasets and train dynamic contextualized word embeddings
by adding a two-layer feed-forward network on top of BERT\textsubscript{BASE} (uncased) and finetuning it for the task of sentiment classification.\footnote{We finetune directly on sentiment analysis without prior finetuning on masked language modeling.}  We again compare against contextualized word embeddings, specifically BERT\textsubscript{BASE} (uncased), which is finetuned without the dynamic component. See Appendix \ref{app:hyper-2} for details 
about implementation, hyperparameter tuning, and runtime.

Dynamic contextualized word embeddings achieve slight but significant
improvements over the already strong performance of non-dynamic BERT  (Table \ref{tab:sentiment-analysis}).\footnote{Statistical significance is tested with a McNemar's test for binary data \citep{McNemar.1947, Dror.2018}.} This provides further
evidence that infusing social and temporal information on the lexical
level can be useful for NLP tasks.

\begin{table} [t!]\centering
\resizebox{0.5\linewidth}{!}{%
\begin{tabular}{@{}lrrrr@{}}
\toprule
{}  & \multicolumn{2}{c}{ Ciao} & \multicolumn{2}{c}{ YELP}\\
\cmidrule(lr){2-3}
\cmidrule(l){4-5}
Model & Dev & Test & Dev & Test   \\
\midrule
DCWE & \best{{\underline{.894}}} & \best{{\underline{.896}}} & \best{{\underline{.969}}} &  \best{{\underline{.968}}} \\
CWE & .889 & .890 & .967 & .966 \\ 
\bottomrule
\end{tabular}
}
\caption{F1 score on sentiment
analysis (higher is better). DCWE:
dynamic contextualized word embeddings; CWE: contextualized word embeddings. The better score per column (highlighted in gray)
is underlined if it is significantly ($p < .01$) better as shown by a McNemar's test for binary data.}  \label{tab:sentiment-analysis}
\end{table}

\section{Conclusion}

We have introduced dynamic contextualized
word embeddings that represent words as a function of both linguistic
and extralinguistic context. Based on a PLM, specifically BERT, 
dynamic contextualized word embeddings model time and social space jointly, which 
makes them advantageous for various areas in NLP. We have trained dynamic contextualized word embeddings on
four datasets and showed that they are capable of tracking social and temporal variability in word meaning. Besides serving as an analytical tool, 
dynamic contextualized word embeddings can also be of benefit for downstream tasks such as sentiment analysis.

\section*{Acknowledgements}

This work was funded by
the European Research Council (\#740516) as well as the Engineering and Physical Sciences Research Council
(EP/T023333/1). 
The first author was also supported by the German Academic Scholarship Foundation and the Arts and Humanities Research Council. 
We thank the anonymous reviewers for
their detailed and extremely helpful comments.

\bibliography{acl2021}
\bibliographystyle{acl_natbib}

\appendix

\section{Appendices}

\subsection{Data Preprocessing} \label{app:preprocessing}

For each dataset, we remove duplicates 
as well as texts with less than 10 words. For the Ciao dataset,
we further remove reviews rated as not helpful.
We lowercase all words. Since BERT's input
is limited to 512 tokens, we truncate longer 
texts by taking the first and last 256 tokens.

\subsection{Embedding Training: Hyperparameters} \label{app:hyper}

\textbf{DCWE.} The hyperparameters of the contextualizer are as for BERT\textsubscript{BASE} (uncased). In particular, 
the dimensionality of the input embeddings $\tilde{ \mathbf{e}}^{(k)}$ is 768. For the dynamic component, the social vectors $\mathbf{s}_{ij}$ and $\tilde{ \mathbf{s} }_i$
have a dimensionality of 50. The node2vec
vectors for the initialization of $\tilde{ \mathbf{s} }_i$ are trained on 10 sampled walks of length 80 per node with a window size of 2. The GAT has two layers with four attention heads, respectively (activation function: $\tanh$). The feed-forward network has two layers (activation function: $\tanh$). We apply dropout with a rate of 0.2 after each layer of the dynamic component. The number of trainable parameters varies between models trained on different datasets due to differences in $|\mathcal{T}|$ and 
is 134,914,570 for ArXiv, 124,990,698 for Ciao, 120,028,762 for Reddit, and 122,509,730 for YELP. We use a batch size of 4 and perform grid search for the number of epochs $n_e \in \{1, \dots ,7\}$, the learning rate $l \in \{ \num{1e-6}, \num{3e-6} \}$,
and the regularization constant $\lambda_a \in \{ \num{1e-2}, \num{1e-1} \}$, thereby
also
determining $\lambda_w$ (Section \ref{sec:training}).

\textbf{CWE.} All hyperparameters are as for BERT\textsubscript{BASE} (uncased). The number of trainable parameters is 110,104,890. We use a batch size of 4 and perform grid search for the number of epochs $n_e \in \{1, \dots ,7\}$ and the learning rate $l \in \{ \num{1e-6}, \num{3e-6} \}$.

For both DCWE and CWE, we tune hyperparameters except for the number of epochs on the Ciao dataset (selection criterion:
masked language modeling perplexity) and use the best  
configuration for ArXiv, Reddit, and YELP. Models are trained with categorical cross-entropy as the loss function and Adam \citep{Kingma.2015} as the optimizer. Experiments are performed on a GeForce GTX 1080 Ti GPU (11GB).

Table \ref{tab:stats-embs} lists statistics of the validation performance over hyperparameter search trials and provides information about best hyperparameter configurations.\footnote{Since expected validation performance \citep{Dodge.2019} may not be correct for grid search, we report mean and standard deviation of the performance instead.} We also report the number of hyperparameter search trials as 
well as runtimes for the hyperparameter search.

\begin{table*} [t!]\centering
\resizebox{\linewidth}{!}{%
\begin{tabular}{@{}lrrrrrrrrrrrrrrrrrrrrrrrr@{}}
\toprule
{} & \multicolumn{6}{c}{ArXiv}  & \multicolumn{6}{c}{ Ciao} & \multicolumn{6}{c}{Reddit}  & \multicolumn{6}{c}{ YELP}\\
\cmidrule(lr){2-7}
\cmidrule(lr){8-13}
\cmidrule(lr){14-19}
\cmidrule(l){20-25}
Model & $\mu$ & $\sigma$ & $n_e$ & $l$  & $\lambda_a$ & $\tau$ & $\mu$ & $\sigma$ & $n_e$ & $l$  & $\lambda_a$ & $\tau$ & $\mu$ & $\sigma$ & $n_e$ & $l$  & $\lambda_a$ & $\tau$ & $\mu$ & $\sigma$ & $n_e$ & $l$  & $\lambda_a$ & $\tau$\\
\midrule
DCWE  & 3.848  & .307 & 7 &3e-6 &1e-1 & 6,756& 6.794 & .606 & 7 & 3e-6 &1e-1 & 11,831 &9.836 & .318 &7 & 3e-6&1e-1 & 4,629 & 5.122 & .384 & 7 & 3e-6&1e-1 & 7,002 \\
CWE   & 3.851 & .305&7 & 3e-6& --- & 3,749& 6.789& .589 &7 & 3e-6& --- & 3,564 & 9.869 & .274 & 7 & 3e-6& --- & 2,160 &5.129 &.384 & 7&3e-6 & --- & 3,551\\ 
\bottomrule
\end{tabular}
}
\caption{Validation performance statistics and hyperparameter search details for embedding training. DCWE:
dynamic contextualized word embeddings; CWE: contextualized word embeddings. The table shows the mean ($\mu$) and standard deviation ($\sigma$) of the validation performance (masked language modeling perplexity) on all hyperparameter search trials and gives the number of epochs ($n_e$),
learning rate ($l$), and regularization constant ($\lambda_a$) with the best
validation performance as well as the runtime ($\tau$) in minutes for one full hyperparameter search (28 trials for DCWE on Ciao, 14 trials for CWE on Ciao, 7 trials for DCWE and CWE on ArXiv, Reddit, and YELP).}  \label{tab:stats-embs}
\end{table*}

\begin{table*} [t!]\centering
\resizebox{\linewidth}{!}{%
\begin{tabular}{@{}lrrrrrrrrrrrrrrrrrrrrrrrr@{}}
\toprule
{} & \multicolumn{6}{c}{ArXiv}  & \multicolumn{6}{c}{ Ciao} & \multicolumn{6}{c}{Reddit}  & \multicolumn{6}{c}{ YELP}\\
\cmidrule(lr){2-7}
\cmidrule(lr){8-13}
\cmidrule(lr){14-19}
\cmidrule(l){20-25}
Model & $\mu$ & $\sigma$ & $n_e$ & $l$  & $\lambda_a$ & $\tau$ & $\mu$ & $\sigma$ & $n_e$ & $l$  & $\lambda_a$ & $\tau$ & $\mu$ & $\sigma$ & $n_e$ & $l$  & $\lambda_a$ & $\tau$ & $\mu$ & $\sigma$ & $n_e$ & $l$  & $\lambda_a$ & $\tau$\\
\midrule
SA  & 3.849 &.302 &7 &3e-6 &1e-1 & 4,438& 6.790 & .635 & 7 &3e-6 &1e-1 & 7,616 &9.851 & .282 &6 &3e-6 &1e-1 & 2,699 &5.127 &.392 & 7& 3e-6&1e-1 &4,231  \\
TA   & 3.860 &.303 & 7 &3e-6 &1e-1 &6,080 & 6.843& .782 & 7&3e-6 &1e-1 & 10,343 &9.871 & .321& 7 &3e-6 &1e-1 &3,859  &5.129 & .388& 7& 3e-6&1e-1 & 6,471 \\ 
\bottomrule
\end{tabular}
}
\caption{Validation performance statistics and hyperparameter search details for ablation study. SA: social ablation; TA: temporal ablation. The table shows the mean ($\mu$) and standard deviation ($\sigma$) of the validation performance (masked language modeling perplexity) on all hyperparameter search trials and gives the number of epochs ($n_e$),
learning rate ($l$), and regularization constant ($\lambda_a$) with the best
validation performance as well as the runtime ($\tau$) in minutes for one full hyperparameter search (28 trials on Ciao, 7 trials on ArXiv, Reddit, and YELP).}  \label{tab:stats-abl}
\end{table*}

\begin{table} [t!]\centering
\resizebox{\linewidth}{!}{%
\begin{tabular}{@{}lrrrrrrrrrrrr@{}}
\toprule
{}  & \multicolumn{6}{c}{ Ciao}   & \multicolumn{6}{c}{ YELP}\\
\cmidrule(lr){2-7}
\cmidrule(l){8-13}
Model & $\mu$ & $\sigma$ & $n_e$ & $l$  & $\lambda_a$ & $\tau$ & $\mu$ & $\sigma$ & $n_e$ & $l$  & $\lambda_a$ & $\tau$ \\
\midrule
DCWE  & .883 & .010 &4 & 3e-6&1e-1 & 8,128 &.967 & .003& 2&3e-6 &1e-1 & 4,373  \\
CWE   & .880 & .011 & 5 & 3e-6& --- & 2,122 & .967 & .001 & 3 & 3e-6& --- & 2,221 \\ 
\bottomrule
\end{tabular}
}
\caption{Validation performance statistics and hyperparameter search details for sentiment analysis. DCWE:
dynamic contextualized word embeddings; CWE: contextualized word embeddings. The table shows the mean ($\mu$) and standard deviation ($\sigma$) of the validation performance (F1 score) on all hyperparameter search trials and gives the number of epochs ($n_e$),
learning rate ($l$), and regularization constant ($\lambda_a$) with the best
validation performance as well as the runtime ($\tau$) in minutes for one full hyperparameter search (20 trials for DCWE on Ciao, 10 trials for CWE on Ciao, 5 trials for DCWE and CWE on YELP).}  \label{tab:stats-sa}
\end{table}

\subsection{Ablation Study: Hyperparameters} \label{app:hyper-abl}

\textbf{SA.} Words are mapped to offsets using time-specific two-layer feed-forward networks
(activation function: $\tanh$). Both layers have a dimensionality of 768. All other hyperparameters
are as for DCWE with a full dynamic component (Appendix~\ref{app:hyper}). The number of trainable parameters again varies between models trained on different datasets due to differences in $|\mathcal{T}|$ and 
is 133,728,570 for ArXiv, 124,279,098 for Ciao, 119,554,362 for Reddit, and 121,916,730 for YELP. We use a batch size of 4 and perform grid search for the number of epochs $n_e \in \{1, \dots ,7\}$, the learning rate $l \in \{ \num{1e-6}, \num{3e-6} \}$,
and the regularization constant $\lambda_a \in \{ \num{1e-2}, \num{1e-1} \}$, thereby
also
determining $\lambda_w$ (Section \ref{sec:training}).

\textbf{TA.} All hyperparameters
are as for DCWE with a full dynamic component (Appendix \ref{app:hyper}), with the difference that we only use one 
social component (consisting of a two-layer GAT and a two-layer feed-forward network) for all time units.
The number of trainable parameters is 111,345,374. We use a batch size of 4 and perform grid search for the number of epochs $n_e \in \{1, \dots ,7\}$, the learning rate $l \in \{ \num{1e-6}, \num{3e-6} \}$,
and the regularization constant $\lambda_a \in \{ \num{1e-2}, \num{1e-1} \}$.

For both SA and TA, we tune hyperparameters except for the number of epochs on the Ciao dataset (selection criterion:
masked language modeling perplexity) and use the best  
configuration for ArXiv, Reddit, and YELP. Models are trained with categorical cross-entropy as the loss function and Adam as the optimizer. Experiments are performed on a GeForce GTX 1080 Ti GPU (11GB).

Table \ref{tab:stats-abl} lists statistics of the validation performance over hyperparameter search trials and provides information about best hyperparameter configurations. We also report the number of hyperparameter search trials as 
well as runtimes for the hyperparameter search.

\subsection{Sentiment Analysis: Hyperparameters} \label{app:hyper-2}

\textbf{DCWE.} The mid layer of the feed-forward network on top of BERT has a dimensionality of 100.
All other hyperparameters are as for DCWE trained on masked language modeling (Appendix~\ref{app:hyper}).  The number of trainable parameters again varies between models trained on different datasets due to differences in $|\mathcal{T}|$ and 
is 124,445,049 for Ciao and 121,964,081 for YELP. We use a batch size of 4 and perform grid search for the number of epochs $n_e \in \{1, \dots ,5\}$, the learning rate $l \in \{ \num{1e-6}, \num{3e-6} \}$,
and the regularization constant $\lambda_a \in \{ \num{1e-2}, \num{1e-1} \}$, thereby
also
determining $\lambda_w$ (Section \ref{sec:training}).

\textbf{CWE.} The mid layer of the feed-forward network on top of BERT has a dimensionality of 100. All other hyperparameters are as for BERT\textsubscript{BASE} (uncased). The number of trainable parameters is 109,559,241. We use a batch size of 4 and perform grid search for the number of epochs $n_e \in \{1, \dots ,5\}$ and the learning rate $l \in \{ \num{1e-6}, \num{3e-6} \}$.

For both DCWE and CWE, we tune hyperparameters except for the number of epochs on the Ciao dataset (selection criterion:
F1 score) and use the best  
configuration for YELP. Models are trained with binary cross-entropy as the loss function and Adam as the optimizer. Experiments are performed on a GeForce GTX 1080 Ti GPU (11GB).

Table \ref{tab:stats-sa} lists statistics of the validation performance over hyperparameter search trials and provides information about best hyperparameter configurations. We also report the number of hyperparameter search trials as 
well as runtimes for the hyperparameter search.

\end{document}